\pdfoutput=1

\documentclass[11pt]{article}

\usepackage[]{acl}
\usepackage{booktabs} 
\usepackage{multirow} 
\usepackage{graphicx} 
\usepackage{amsmath}
\usepackage{times}
\usepackage{latexsym}
\usepackage{color, soul}
\usepackage{placeins}
\usepackage{float}

\newcommand{\tss}[1]{\textsubscript{\scriptsize$\pm$#1}}

\usepackage[T1]{fontenc}

\usepackage[utf8]{inputenc}

\usepackage{microtype}

\usepackage{inconsolata}

\usepackage{subcaption}
\usepackage{tikz}

\title{A Multi-Dimensional Evaluation of Explainability in Media Bias Detection}

\author{
Ting Chen$^{1, 2}$
\quad Raina Zhang$^{2}$ 
\quad Benjamin M. Ampel$^{3}$ 
\quad Sagar Samtani$^{2}$
\\
$^1$Carnegie Mellon University
\quad $^2$Indiana University
\quad $^3$Georgia State University\\
 \texttt{tingc2@andrew.cmu.edu}
 \quad \texttt{bampel@gsu.edu}
 \quad \texttt{\{tch4, cz1, ssamtani\}@iu.edu}
}

\begin{document}
\maketitle

\begin{abstract}
Detecting media bias automatically is difficult because biased framing is often subtle, yet in domains such as news analysis, accurate predictions alone are insufficient without explanations that reflect the model's underlying reasoning. We present a multi-dimensional evaluation of explainability in encoder-based media bias detection using the Bias Annotations By Experts (BABE) dataset. Specifically, we study BERT and RoBERTa as classifiers (base and large variants) along three complementary axes: predictive performance, explanation plausibility (token-level alignment with expert rationales), and mechanistic faithfulness (whether compact sets of attention heads recover predictive signal under counterfactual rationale masking). To induce variation in plausibility, we additionally investigate attention-supervised finetuning, which incorporates expert rationale annotations as an auxiliary training signal. Attention supervision serves as an intervention on attribution plausibility, while the effectiveness of attribution methods varies substantially across architectures. Circuit analysis further reveals substantial variation in mechanistic recoverability across architectures, suggesting that model scale alone does not determine circuit compressibility. Taken together, our findings suggest that predictive performance, attribution plausibility, and mechanistic faithfulness characterize different aspects of model behavior and should be evaluated separately when studying explainability in media bias detection.
\end{abstract}

\section{Introduction}

As news consumption moves online, the scale of media production has outpaced the capacity for manual bias analysis. Early detection methods relied on content analysis and qualitative frameworks from the social sciences \cite{hamborg_automated_2019}, but these approaches lack scalability and suffer from annotation inconsistencies. Transformer-based models now surpass manual annotation in consistency \cite{spinde_media_2024}, and large language models (LLMs) can capture the linguistic nuance through which bias is expressed \cite{wang_adaptable_2024}.

However, strong classification performance alone is insufficient for media bias detection, where predictions carry editorial and civic weight. A system that flags an article as biased without interpretable reasoning may be dismissed by journalists or misapplied by policymakers \cite{lyu_towards_2024}.

One of the main concerns in eXplainable AI (XAI) is that two key dimensions are often conflated but should be kept separate \citep{lyu_towards_2024}. Plausibility measures whether an explanation aligns with human judgment (i.e., do the highlighted tokens match what an expert annotator would select?) Faithfulness measures whether an explanation reflects the model's actual computation (i.e., does the model rely on those tokens to reach its prediction?) A model can produce plausible but unfaithful explanations when the XAI mechanism concentrates on semantically reasonable tokens that play no causal role in the prediction. Faithful explanations may highlight tokens that a human would not intuitively select. Conflating the two can lead to overconfident assessments of the quality of the explanation.

Prior work on explainable bias detection has focused almost entirely on plausibility, lacking consideration on model faithfulness. For example, attention-based XAI methods are widely implemented \citep{zheng_attention_2025}, but whether attention constitutes a faithful explanation has long been contested \citep{jain2019attentionexplanation}. Raw attention scores often assign the highest weights to function words and delimiters, undermining their reliability even when the overall distribution appears plausible. 

Mechanistic interpretability provides a complementary perspective to attention-based plausibility evaluation by focusing on the internal computations responsible for a model's predictions. Circuit discovery methods, such as ACDC \citep{conmy2023automatedcircuitdiscoverymechanistic}, identify the minimal computational subgraphs causally responsible for model behavior. However, current work has focused almost exclusively on decoder-only language models, and the relationship between mechanistic faithfulness and attribution plausibility remains largely unexplored for encoder-based text classification.

We bridge this gap by jointly evaluating plausibility and faithfulness for encoder-based media bias classifiers. Our contributions are:

\begin{enumerate}
    \item We conduct a multi-dimensional evaluation of explainability in media bias detection, jointly studying predictive performance, attribution plausibility, and mechanistic faithfulness.
    
    \item We adapt activation-patching circuit discovery to encoder-only text classification, introducing Retention and Rescue metrics to evaluate circuit recoverability under counterfactual rationale masking.
    
    \item Using attention supervision as an intervention on plausibility, we characterize how the quality of the explanation varies between attribution methods, model architectures, and mechanistic circuit properties.
\end{enumerate}

\section{Related Work}

\paragraph{Automated Bias Detection.}
Early work on media bias detection relied primarily on handcrafted linguistic and contextual features. Previous studies have shown that lexical framing and contextual cues can effectively identify bias-inducing language \citep{SPINDE2021102505}. More recent work shifted to transformer-based architectures, where studies on the BABE benchmark \citep{spinde_neural_2021} demonstrated that contextualized transformer models substantially outperform traditional feature-based approaches in fine-grained bias classification tasks. 

Recent research has also explored the use of LLMs for political bias analysis. GPT models can achieve high agreement with human political bias ratings even in zero-shot settings \citep{hernandes2024llmsleftrightcenter}. Despite the strong performance of LLMs, concerns also arose. Studies have demonstrated that larger language models may still exhibit systematic political framing biases despite improved generation quality \citep{huang2026biggerisntbettercomprehensive}. Together, these findings motivate the need for more transparent and faithful explainability methods for bias detection systems.

\paragraph{Explainable NLP.}
Explainability methods in NLP span post-hoc feature attribution, gradient-based analysis, attention interpretation, probing, and mechanistic circuit analysis \citep{lyu_towards_2024, luo_understanding_2024, tursunalieva_making_2024}. The breadth of available methods makes multidimensional evaluation necessary.

Rationale-based evaluation frameworks, such as ERASER \citep{deyoung_eraser_2020}, formalize the assessment of explanations using human-annotated spans. Chain-of-thought prompting \citep{yeo_how_2024, wei_chain--thought_2023} takes a different approach by eliciting intermediate reasoning steps from the model itself.

Attention-based explanations are among the most studied. \citeauthor{hao_self-attention_2021} and \citeauthor{naim_explaining_2024} explore the conditions under which attention provides interpretable insights into the information flow of the transformer. More recent work improves the explainability of self-attention distributions in text classification \citep{mylonas_improving_2022}, and studies demonstrate that specific attention heads correspond to semantically meaningful reasoning patterns \citep{zheng_attention_2025}. However, whether attention scores constitute genuine explanations remains debated \cite{wiegreffe_attention_2019, jain2019attentionexplanation}.

To move beyond the limitations of attention analysis, the mechanistic interpretability shifts the focus from individual weights to localized computational subgraphs known as circuits \citep{olsson2022incontextlearninginductionheads, elhage2021mathematical}. This line of work has focused primarily on autoregressive, decoder-only architectures. A prominent example is the indirect object identification circuit (IOI) in GPT-2, where activation patching revealed how heads cooperatively route linguistic information to produce predictions \citep{conmy2023automatedcircuitdiscoverymechanistic}. More recent work has further emphasized the evaluation of circuit discovery methods through behavioral recovery rather than qualitative inspection alone, benchmarking methods by how well recovered circuits preserve model behavior \citep{syed-etal-2024-attribution, arad-etal-2025-findings}. However, the application of circuit discovery to encoder-based text classification and the comparison of the resulting faithfulness measures against plausibility scores on the same models have not been explored.

Incorporating human rationales during finetuning has also been used to shape model behavior beyond raw classification accuracy, including improving reasoning and reducing hallucination in other settings \citep{just_data-centric_2025, cho-etal-2025-mechanistic}.

\section{Method}

\begin{figure*}[t]
    \centering
    \includegraphics[width=\textwidth]{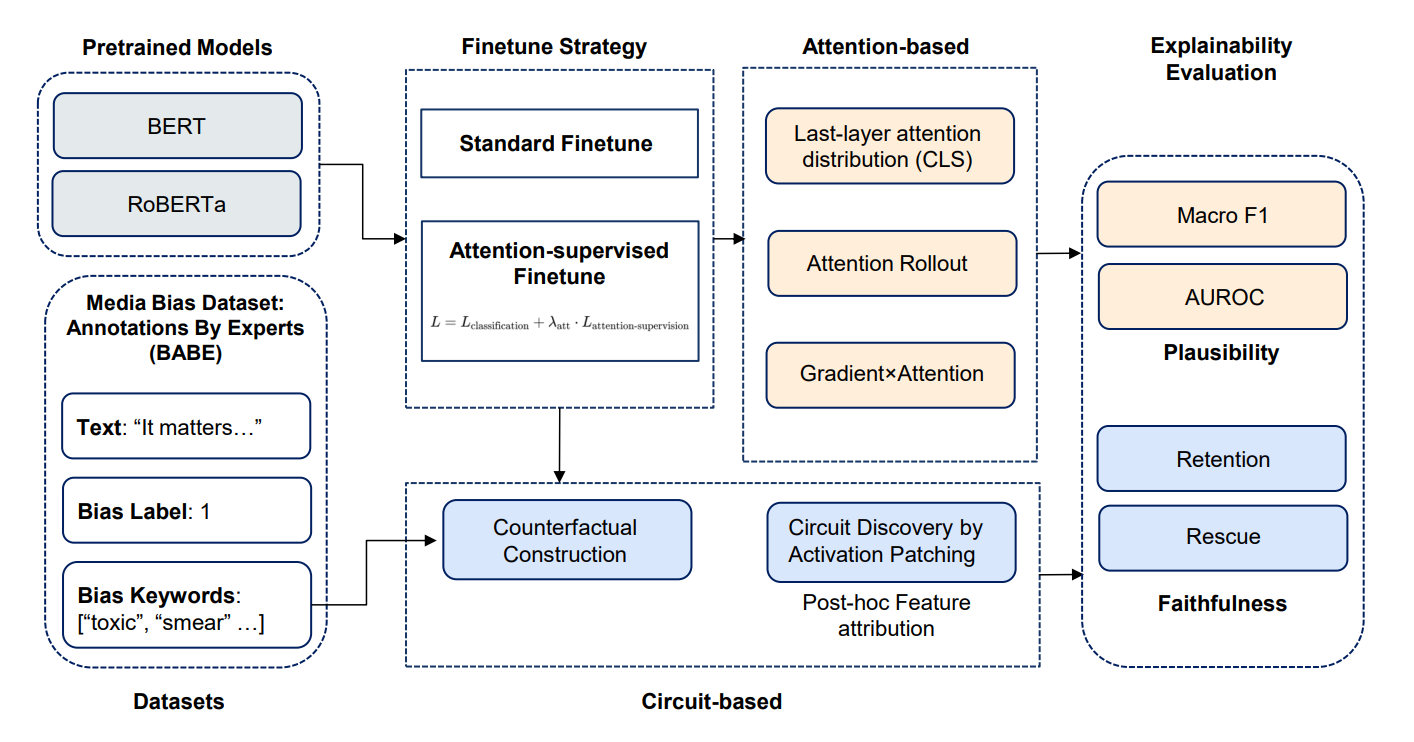}
    \caption{Flowchart of the media bias analysis pipeline.}
    \label{fig:bias-flowchart}
\end{figure*}

Our study progresses in three stages, illustrated in Figure~\ref{fig:bias-flowchart}. First, we train the BERT and RoBERTa classifiers on the BABE dataset under both standard supervision and attention-supervised finetuning, using rationale supervision as a controlled intervention on explanation plausibility. Second, we evaluated token-level plausibility using three attention-based attribution methods and compared their outputs against expert rationale annotations. Third, we evaluate mechanistic faithfulness using activation-patching circuit discovery, measuring the extent to which compact sets of attention heads recover predictive signal under counterfactual rationale masking.

\subsection{Dataset}

\paragraph{Bias Annotations By Experts (BABE).}
We use the BABE dataset \citep{spinde_neural_2021}, which contains 3,700 English news sentences drawn from 14 U.S.\ outlets spanning left, center, and right political leanings according to the AllSides media bias chart. Each sentence carries a binary bias label assigned by majority vote among five trained annotators with backgrounds in data science, psychology, and intercultural communication (Krippendorff's $\alpha = 0.40$ for the bias label). We downsample the majority class to produce a balanced subset of 2,762 sentences (1,381 biased, 1,381 unbiased). For biased instances, the annotators additionally highlighted specific words and phrases responsible for the biased framing. We use these token-level rationales both as supervision during attention-guided fine-tuning and as ground truth for evaluating the plausibility of explanations. We partition the data using a stratified 70:15:15 train, validation, and test split to ensure class balance across all subsets.

\subsection{Models}

We experiment with two encoder-only transformer architectures: BERT and RoBERTa. BERT \citep{devlin2019bertpretrainingdeepbidirectional} uses bidirectional masked language modeling to produce contextual token representations; its \texttt{[CLS]} embedding serves as input for sequence-level bias classification. RoBERTa \citep{liu2019robertarobustlyoptimizedbert} modifies BERT's pretraining procedure with dynamic masking, larger batches, and more data, yielding stronger downstream performance on tasks that require sensitivity to subtle lexical choices. For both architectures, we experiment with base and large variants to examine how model capacity affects classification performance, plausibility, and faithfulness.

\subsection{Attention-supervised Finetuning}

Expert rationales in BABE make it possible to train the model on both bias labels and on \emph{why} a sentence is biased. We incorporate an auxiliary attention supervision loss \cite{deyoung_eraser_2020, cho-etal-2025-mechanistic} that encourages the model to allocate higher attention mass to expert-highlighted tokens. We use rationale supervision as an intervention on attribution plausibility. The joint training objective is:
\[
\mathcal{L} = \mathcal{L}_{\text{cls}} + \lambda \cdot \mathcal{L}_{\text{attn}},
\]
where $\mathcal{L}_{\text{cls}}$ is the binary cross-entropy loss over the bias label, $\mathcal{L}_{\text{attn}}$ is the mean squared error between the normalized last-layer CLS attention distribution and the binary expert rationale mask:
\[
\mathcal{L}_{\text{attn}} = \frac{1}{N} \sum_{i=1}^{N} \left( \bar{a}_i - r_i \right)^2,
\]
where $\bar{a}_i$ denotes the min-max normalized attention weight from the CLS token to the token $i$ and $r_i \in \{0, 1\}$ is the expert rationale indicator. $\lambda$ controls the tradeoff between predictive accuracy and rationale alignment. We fix $\lambda = 0.5$ across all experiments based on a hyperparameter search ($\lambda \in \{0, 0.2, 0.5, 0.7, 1.0, 1.5, 2.0, 5.0\}$).

\subsection{Attention-based Explainability}

\paragraph{Last-layer CLS Attention.}
In the final layers of BERT, a large part of the attention heads focus heavily on task-related tokens \citep{clark2019doesbertlookat}. The most straightforward attribution approach examines the attention distribution from the \texttt{[CLS]} token at the final transformer layer $L$. The importance score for token $i$ is:
\[
s_i^{\text{CLS}} = \text{A}^{(L)}_{\text{CLS},\, i}.
\]

where $\text{A}^{(L)}_{\text{CLS},\, i}$ is the weight of attention from 
the \texttt{[CLS]} token to token $i$ in the final transformer layer
$L$. This formulation treats higher attention as greater importance to
the prediction, but ignores how information is routed through earlier
layers and may overweight tokens that are prominent only at the final
layer.

\paragraph{Attention Rollout.}
Attention rollout \citep{abnar2020quantifyingattentionflowtransformers} addresses the limitation of examining only the last single layer by propagating attention across all layers via recursive matrix multiplication, approximating the flow of token information from input to output. At each layer $l$, we first average the attention matrices across all heads, then add the identity matrix to account for the residual connection:
\[
\tilde{\text{A}}^{(l)} = 0.5 \cdot \bar{\text{A}}^{(l)} + 0.5 \cdot I,
\]
where $\bar{\text{A}}^{(l)}$ is the head-averaged attention matrix at layer $l$. The rollout attribution is then computed as follows:
\begin{align*}
\text{Rollout}^{(L)} &= \tilde{\text{A}}^{(1)} \tilde{\text{A}}^{(2)}
\cdots \tilde{\text{A}}^{(L)}, \\
s_i^{\text{rollout}} &= \text{Rollout}^{(L)}_{\text{CLS},\, i}.
\end{align*}

\paragraph{Gradient-weighted Attention.}
Raw attention weights indicate where the model looks, but not whether that attention affects the output. Following \citeauthor{hao_self-attention_2021}, we compute gradient-weighted attention (i.e., Gradient$\times$Attention) by scaling the final-layer attention weights by their gradient with respect to the pre-sigmoid logit $y$ for the predicted class using:
\[
s_i^{\text{grad}} = \text{A}^{(L)}_{\text{CLS},\, i} \cdot 
\frac{\partial y}{\partial \text{A}^{(L)}_{\text{CLS},\, i}}.
\]

The gradient term measures how sensitive the prediction is to each attention weight, so the product downweights attention heads that are large but causally inert. The result is an attribution that combines the way the model is attended with how much that attention matters to the classification decision.

\subsection{Circuit Discovery for Explainability}

Attention-based attribution scores measure plausibility, but not whether the model actually relies on the highlighted tokens. To assess faithfulness, we adapt Automated Circuit Discovery (ACDC) \citep{conmy2023automatedcircuitdiscoverymechanistic} to encoder-based classifiers. ACDC identifies the minimal subgraph of a model's computation graph whose activations are sufficient to reproduce the model's predictions.

\paragraph{Counterfactual Masking.} We converted sentence-level samples from the BABE dataset \citep{spinde_media_2024} into counterfactual pairs. Let $\mathcal{D} = \{(T_i, y_i, K_i)\}$ represent the dataset, where $T_i$ is the raw textual string, $y_i \in \{0, 1\}$ is the binary expert bias flag, and $K_i$ is a set of expert-annotated biased keywords or phrases. 

We apply a deterministic corruption function that replaces instances of biased keywords or phrases with an architectural token mask (\texttt{[MASK]}). Counterfactual pairs are partitioned using a stratified $70:15:15$ train, validation, and test split strategy.

\paragraph{Linear Logistic Probing.}
Circuit analysis is performed after fine-tuning at each model checkpoint. Rather than using the architecture-specific classification heads of BERT and RoBERTa as causal targets, we trained a lightweight linear probe over the final hidden representations to provide a uniform diagnostic objective across encoder architectures. This design allows the same activation-patching pipeline to operate directly on the shared transformer encoder layers without introducing architecture-specific branching in the downstream readout.

Let
\[
\mathbf{h}_i^{(L)} = \text{Encoder}(T_i)_{[\text{CLS}]} \in \mathbf{R}^d
\]
denote the final-layer representation of the classification token (\texttt{[CLS]}) for clean input $T_i$. We train a logistic regression classifier on a sampled subset of clean hidden representations to predict the target bias label $y_i$:
\begin{equation}
P(y_i = 1 \mid \mathbf{h}_i^{(L)})
=
\sigma(\mathbf{w}^{T}\mathbf{h}_i^{(L)} + b),
\end{equation}
where $\mathbf{w}\in\mathbf{R}^{d}$ and $b\in\mathbf{R}$ are learned parameters and $\sigma(\cdot)$ denotes the sigmoid activation function. After optimization, the fixed probe parameters $(\mathbf{w}, b)$ serve as the downstream causal readout used during activation patching experiments.

\paragraph{Circuit Extraction by Activation Patching.} To quantify the contribution of each attention head, we perform activation patching on all head positions. For each head $h$ at layer $\ell$, we run a forward pass on the corrupted input but replace that head's activations with the corresponding clean activations from $\mathcal{C}_{\text{clean}}$.

The causal effect $\mathcal{E}_{(\ell, h)}$ of an individual head is defined as the absolute marginal difference in the expected probability output of the downstream probe:
\begin{align*}
    \mathcal{E}_{(\ell, h)} =& \left| \mathbf{E} \left[ P(y=1 \mid \mathbf{h}_{\text{patched}(\ell, h)}^{(L)}) \right] \right. \\
    &- \left. \mathbf{E} \left[ P(y=1 \mid \mathbf{h}_{\text{corrupt}}^{(L)}) \right] \right|
\end{align*}
Taking the absolute value ensures that heads contributing to predictions of either class (biased or unbiased) are ranked by the magnitude of their causal influence rather than direction. Subsequently, all heads are sorted in descending order of $\mathcal{E}_{(\ell, h)}$ to form a causal importance ranking.

\paragraph{Circuit Verification Modes.} To evaluate the collective characteristics of the causal heads ranked in the top-$k$, we assemble them into a candidate circuit sub-network $\mathcal{G}_k = \{(\ell, h)_1, (\ell, h)_2, \dots, (\ell, h)_K\}$. We implement two complementary patch styles via forward hooks to assess the circuit's functional traits on the held-out test split:

\begin{enumerate}
    \item \textbf{Circuit Elimination (\texttt{no} mode):} we run clean inputs $T$ through the model but replace the outputs of all heads in $\mathcal{G}_k$ with their corrupted counterparts from $\mathcal{C}_{\text{corrupt}}$. This tests whether the circuit is \emph{causally necessary} for the model's predictive gap.
    \item \textbf{Circuit Isolation (\texttt{circuit\_only} mode):} We run the corrupted inputs $T^{\text{corrupt}}$ but restore clean activations from $\mathcal{C}_{\text{clean}}$ for the heads in $\mathcal{G}_k$ only. This tests whether the circuit is \emph{causally sufficient} (sufficient to recover the original predictive behavior under activation patching.) to recover the model's predictive signal.
\end{enumerate}

Since circuit size is an important hyperparameter, we sweep $k \in \{1, 3, 5, 10, 20, 30\}$ to characterize the faithfulness-compactness corelation for each model.
\subsection{Evaluation Metrics}

\paragraph{Macro F1.}
We evaluate classification performance using Macro F1, which computes F1 
independently for each class and averages with equal weight. This protects 
against performance inflation from class imbalances and ensures that reported 
gains reflect genuine improvement in both biased and unbiased instances.

\paragraph{AUROC for Attention Plausibility.} The Area Under the Receiver Operating Characteristic Curve (AUROC) is a standard metric for evaluating token-level ranking performance \citep{MANDREKAR20101315, deyoung_eraser_2020}. To evaluate whether attention-based attributions align with human rationales, we computed the AUROC between token-level attribution scores and binary expert rationale masks. An AUROC of 0.5 indicates that the attributions are no better than random when identifying the expert-highlighted tokens; a value of 1.0 indicates perfect alignment. AUROC measures the plausibility of the explanation \emph{plausibility}: how well the model-derived token importance matches the human judgment.

\paragraph{Retention and Rescue for Circuit Faithfulness.} To evaluate circuit faithfulness, we compare the predicted probability gap of the model under three evaluation conditions: the unperturbed baseline model ($\Delta P(\text{base})$), the circuit-eliminated model ($\Delta P(\text{none})$), and the circuit-isolated model ($\Delta P(\text{circuit\_only})$). From these conditions, we adapt the sufficiency and comprehensiveness measures introduced in the ERASER benchmark \cite{deyoung_eraser_2020} and inspired by \citeauthor{conmy2023automatedcircuitdiscoverymechanistic}, reframed here to evaluate causal circuits rather than discrete token rationales. We present two complementary causal metrics, Retention and Rescue. \textbf{Retention} quantifies the fraction of the full model's probability gap that the isolated subgraph reproduces:
\[
\text{retention} = \frac{\Delta P(\text{circuit\_only})}{|\Delta P(\text{base})| + \epsilon},
\]
where $\epsilon = 10^{-12}$ is a regularizing scalar to prevent division by zero. The absolute value in the denominator ensures that retention remains interpretable when the base probability gap is negative (i.e., when the model's mean predicted probability for unbiased instances exceeds that for biased instances). Retention of $1.0$ indicates that the circuit fully reproduces the probability gap of the base model, while values near $0$ indicate that the circuit captures little predictive signal. \textbf{Rescue} measures the signed predictive margin by which the circuit recovers beyond the fully ablated baseline.
\[
\text{rescue} = \Delta P(\text{circuit\_only}) - \Delta P(\text{none}).
\]
Positive rescue indicates that the circuit recovers the predictive signal beyond what the ablated model retains. The rescue score complements retention by confirming that high retention values are not inflated by a near-zero ablated baseline. Together, retention and rescue quantify faithfulness by measuring the degree to which the isolated circuit is causally responsible for the model's bias predictions.
\section{Results}
\label{sec:results}
\begin{table*}[t]
\centering
\small
\resizebox{\textwidth}{!}{
    \begin{tabular}{llcccc}
    \toprule
    \multirow{2}{*}{\textbf{Model}} &
    \multirow{2}{*}{\textbf{Finetune Strategy}} &
    \multirow{2}{*}{\textbf{Macro F1}} &
    \multicolumn{3}{c}{\textbf{Explanation Plausibility (AUROC)}} \\
    \cmidrule(lr){4-6}
    & & & \textbf{CLS} & \textbf{Rollout} & \textbf{Grad $\times$ Attn} \\
    \midrule

    BERT baseline \cite{spinde_neural_2021} & -- & 0.762 & -- & -- & -- \\
    \midrule

    \multirow{2}{*}{BERT (base)} & Standard & 0.829 & 0.523 (0.218) & 0.497 (0.196) & 0.583 (0.297) \\
    & Attention-supervised & 0.805 & 0.476 (0.227) & 0.506 (0.196) & 0.566 (0.284) \\
    \midrule

    \multirow{2}{*}{BERT (large)} & Standard & 0.812 & 0.554 (0.220) & 0.515 (0.162) & 0.621 (0.321) \\
    & Attention-supervised & 0.809 & 0.487 (0.211) & 0.496 (0.170) & 0.535 (0.286) \\
    \midrule

    \multirow{2}{*}{RoBERTa (base)} & Standard & 0.843 & 0.492 (0.283) & 0.426 (0.292) & 0.687 (0.271) \\
    & Attention-supervised & 0.831 & 0.575 (0.320) & 0.465 (0.287) & 0.674 (0.317) \\
    \midrule

    \multirow{2}{*}{RoBERTa (large)} & Standard & 0.849 & 0.470 (0.284) & 0.484 (0.240) & 0.620 (0.271) \\
    & Attention-supervised & \textbf{0.863} & 0.512 (0.312) & 0.522 (0.231) & 0.661 (0.245) \\

    \bottomrule
    \end{tabular}
}
\caption{Classification performance and explanation plausibility. AUROC values (with standard deviations) measure alignment between attribution method outputs and expert rationales in the BABE dataset.}
\label{tab:attn_full_result}
\end{table*}

\subsection{Attention Analysis and Plausibility}

Table~\ref{tab:attn_full_result} reports Macro F1 and AUROC scores for all models. All transformer models substantially outperform the feature-based baseline reported by \citet{spinde_neural_2021} (Macro F1 = 0.511) and the BERT baseline reported in the same work (Macro F1 = 0.762), with every configuration exceeding $0.80$ Macro F1. RoBERTa-large achieved the strongest predictive performance in general, reaching a Macro F1 of $0.863$ under attention-supervised finetuning.

\paragraph{The effect of attention supervision on predictive performance is modest and architecture-dependent.}
Attention supervision does not produce a uniform improvement across architectures. While RoBERTa-large benefits from rationale supervision, other configurations show smaller gains or slight decreases relative to standard finetuning. This suggests that rationale supervision acts as a weak auxiliary signal whose effects depend on the model architecture and capacity rather than fundamentally changing the learned representations.

\paragraph{Plausibility is strongly attribution- and architecture-dependent.}
We evaluate the plausibility of the explanation using AUROC, comparing CLS attention, Attention Rollout, and Gradient$\times$Attention. No single attribution method consistently dominates across architectures. Gradient$\times$Attention generally performs strongly in BERT models, while Attention Rollout and CLS attention exhibit greater variability in architectures and finetuning conditions. These findings suggest that the quality of the attribution depends not only on the explanation method itself but also on the underlying attention structure of the model being analyzed.

Overall, the alignment between model explanations and expert rationales remains moderate. Most configurations achieve AUROC values only modestly above the random baseline of $0.5$, with substantial variance between examples and architectures. These results are consistent with previous work that argued that attention distributions alone do not reliably constitute faithful explanations of model behavior \cite{jain2019attentionexplanation}.

\begin{figure*}[!t]
    \centering

    \begin{subfigure}[t]{0.48\textwidth}
        \centering
        \includegraphics[width=\textwidth]{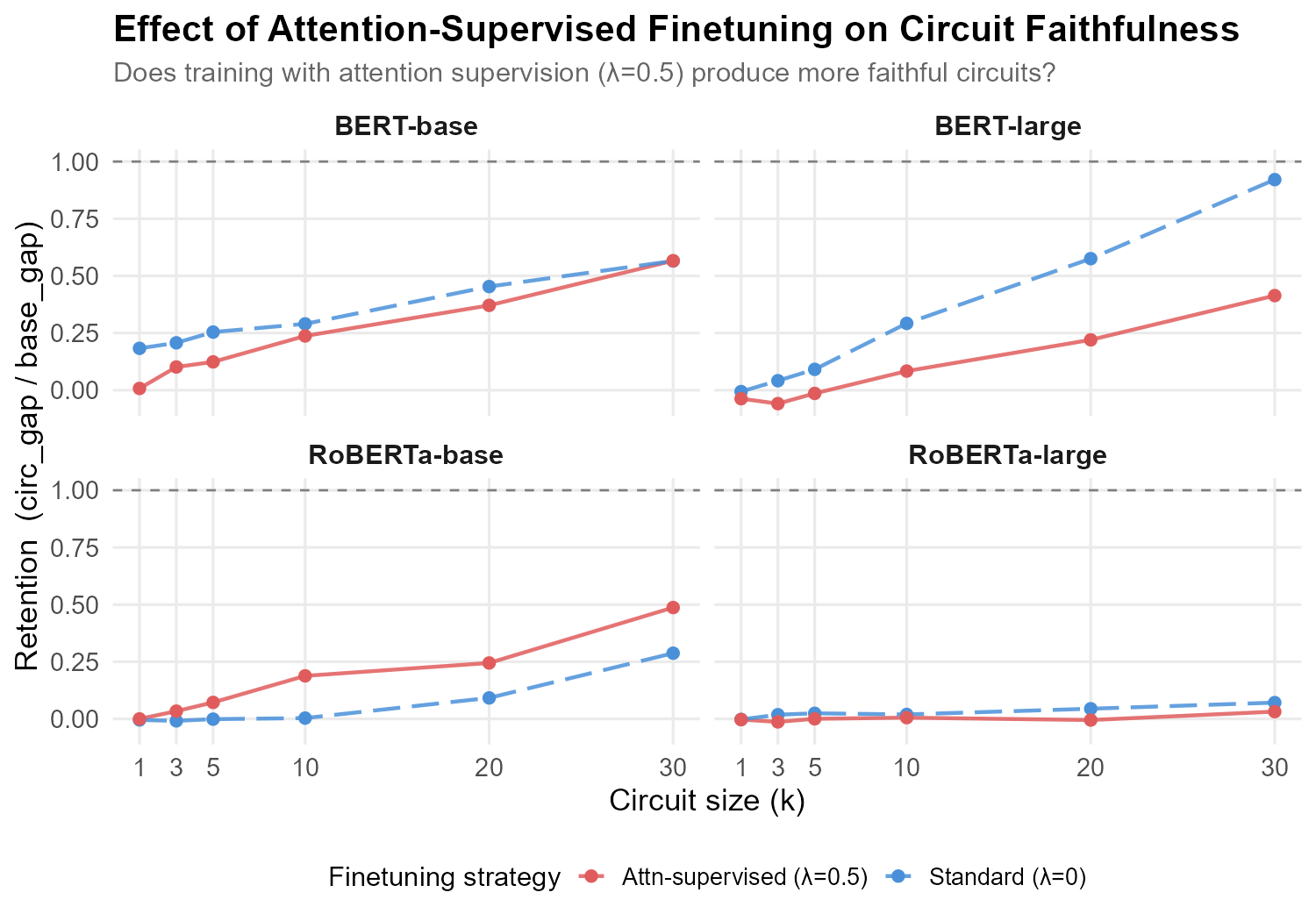}
        \caption{Effect of attention-supervised finetuning.}
        \label{fig:lambda_comp}
    \end{subfigure}
    \hfill
    \begin{subfigure}[t]{0.48\textwidth}
        \centering
        \includegraphics[width=\textwidth]{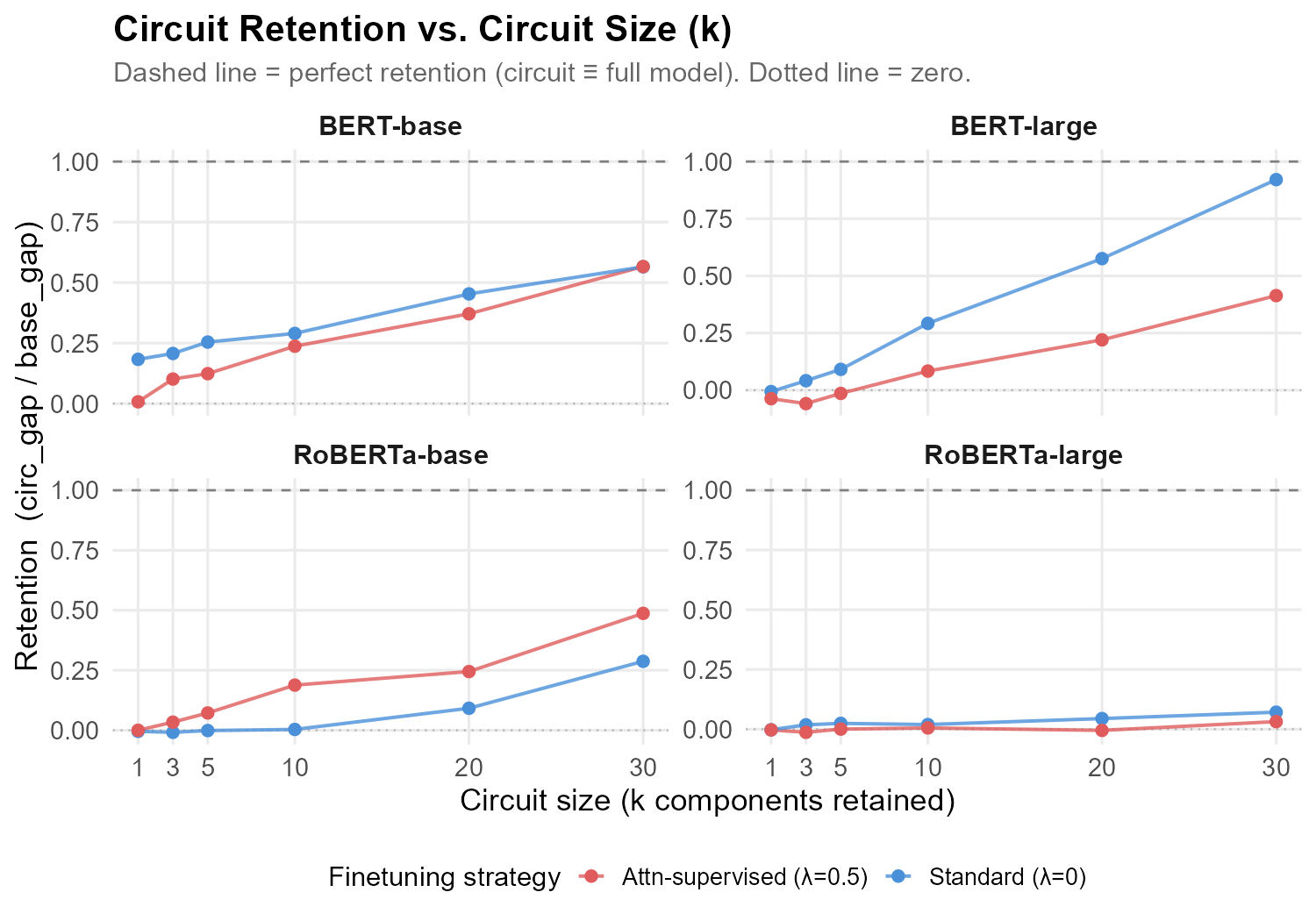}
        \caption{Circuit retention vs. circuit size.}
        \label{fig:retention_by_k}
    \end{subfigure}
    \vspace{0.5em}

    \begin{subfigure}[t]{0.48\textwidth}
        \centering
        \includegraphics[width=\textwidth]{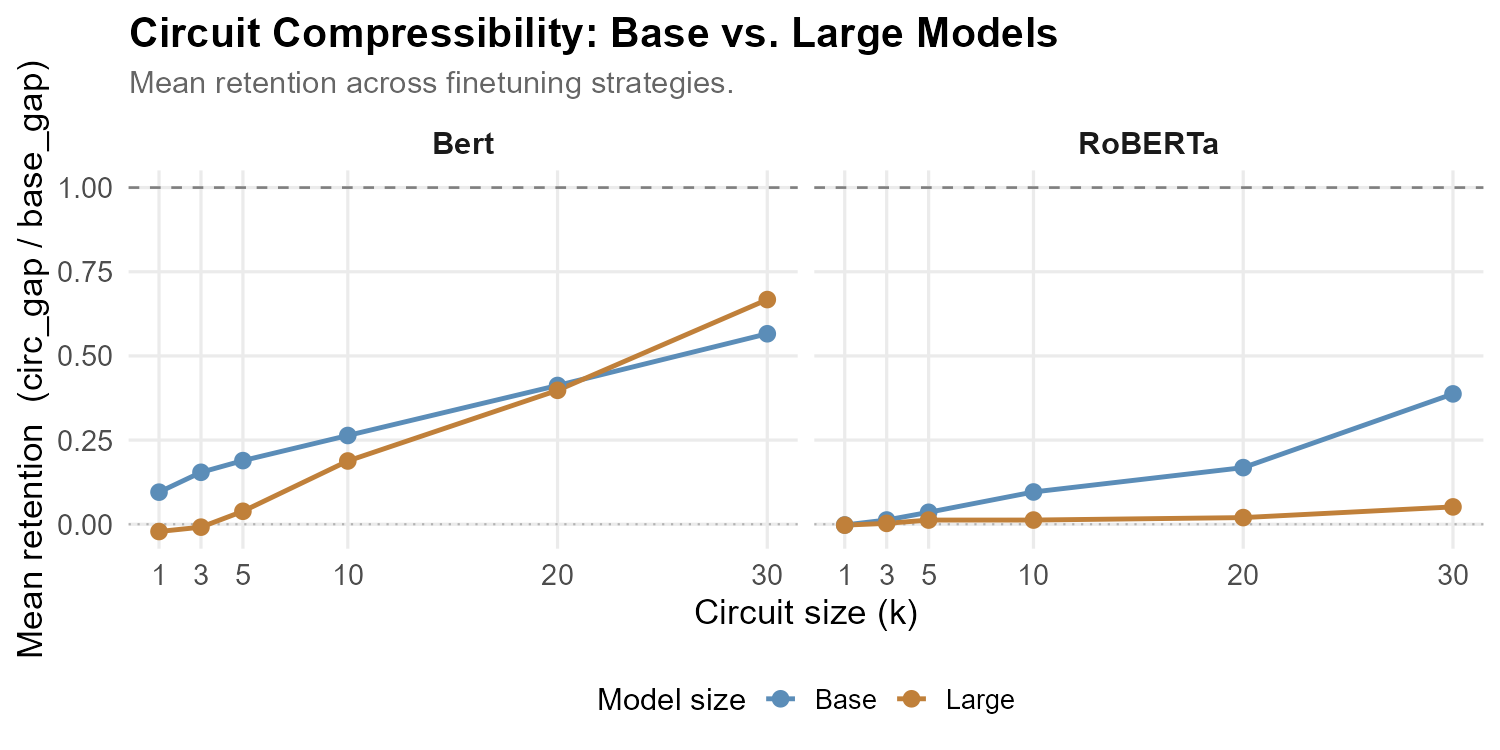}
        \caption{Circuit compressibility across model sizes.}
        \label{fig:base_vs_large}
    \end{subfigure}
    \hfill
    \begin{subfigure}[t]{0.48\textwidth}
        \centering
        \includegraphics[width=\textwidth]{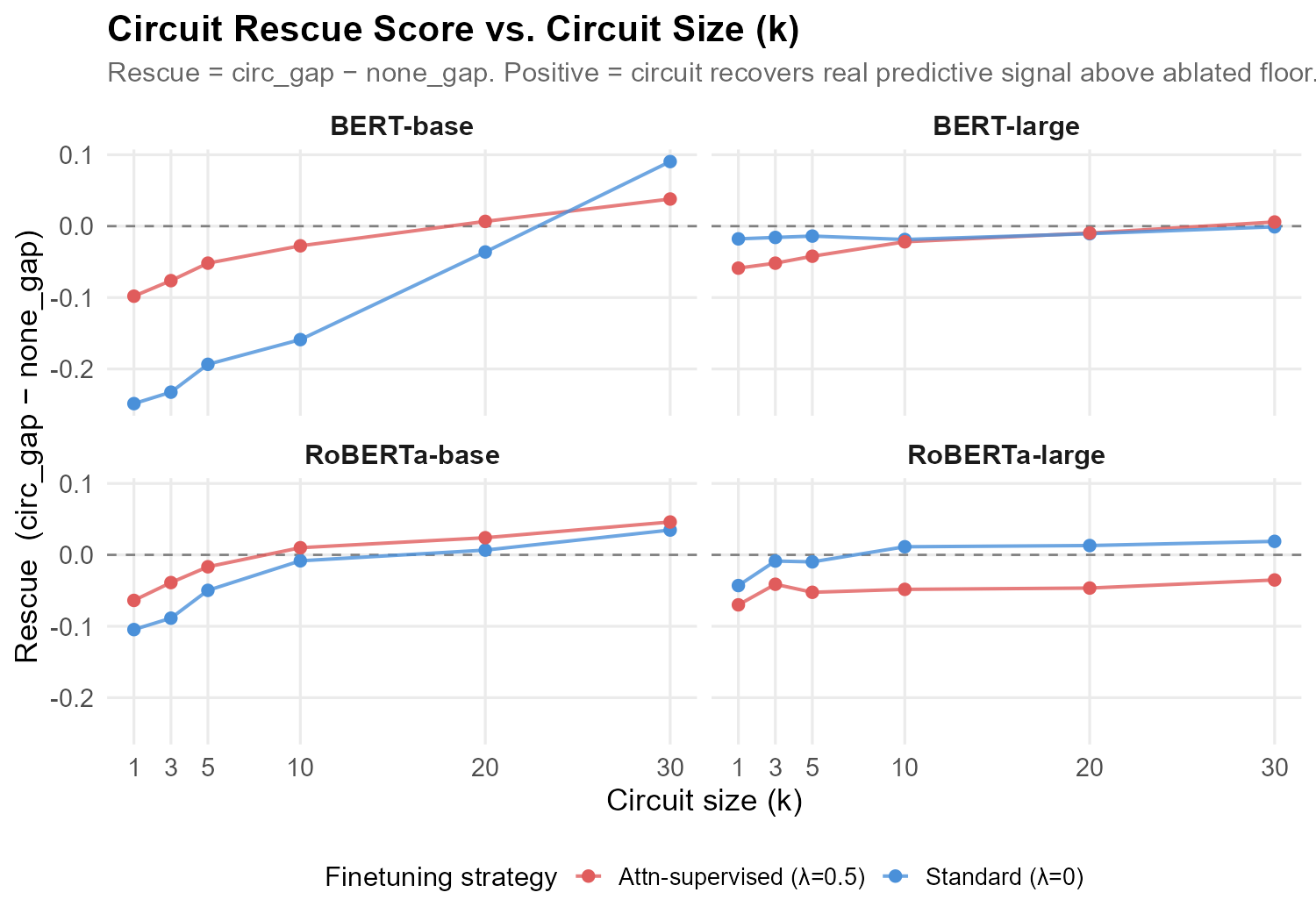}
        \caption{Circuit rescue scores across circuit sizes.}
        \label{fig:rescue}
    \end{subfigure}

    \caption{
    Circuit faithfulness analysis across encoder-based bias detection models.
    Attention-supervised finetuning generally improves circuit retention and rescue scores, particularly in smaller models.
    }
    \label{fig:all_circuit_plots}
\end{figure*}

\subsection{Circuit Analysis and Faithfulness}

\paragraph{Circuit retention varies across architectures and interventions.}
Figure~\ref{fig:lambda_comp} compares retention under standard and attention-supervised finetuning. The effect of rationale supervision is mixed across architectures. The RoBERTa-base exhibits noticeably improved retention under attention supervision, while BERT-large and RoBERTa-large show weaker or even reduced retention relative to standard finetuning. The BERT-base exhibits similar retention under both training strategies once larger circuits are considered. These results suggest that rationale supervision does not uniformly improve mechanistic faithfulness and that circuit recoverability depends strongly on model architecture and representational structure.

\paragraph{Circuit faithfulness varies substantially across architectures.}
Figure~\ref{fig:retention_by_k} shows that retention generally increases with circuit size $k$, indicating that the predictive signal is distributed over multiple components rather than concentrated in a single attention head. However, architectures differ substantially in compressibility (Figure~\ref{fig:base_vs_large}). The BERT-family models exhibit a relatively monotonic increase. The RoBERTa-base shows a similar trend under attention supervision but near-zero retention under standard finetuning, whereas the RoBERTa-large remains difficult to compress even for large circuit sizes. These findings suggest that model architecture and scale influence the degree to which predictive behavior can be recovered using compact mechanistic circuits.

\paragraph{Rescue analysis reveals meaningful causal contribution.}
Retention alone can be inflated when the ablated baseline is near zero, so we also examine the rescue scores (Figure~\ref{fig:rescue}). Positive rescue confirms that the circuit recovers the predictive signal above the ablated baseline. The attention-supervised models of the BERT-base and the RoBERTa-base show increasingly positive rescue as $k$ grows, consistent with the circuits that capture genuine predictive computation. The Standard-finetuned RoBERTa-large, by contrast, remains near or below zero across most settings, pointing to a weak causal structure. Rescue scores become positive only at larger circuit sizes, which indicates that bias detection is not governed by a single sparse mechanism but emerges from coordination among multiple heads.

\begin{figure}[!t]
    \centering
    \includegraphics[width=\columnwidth]{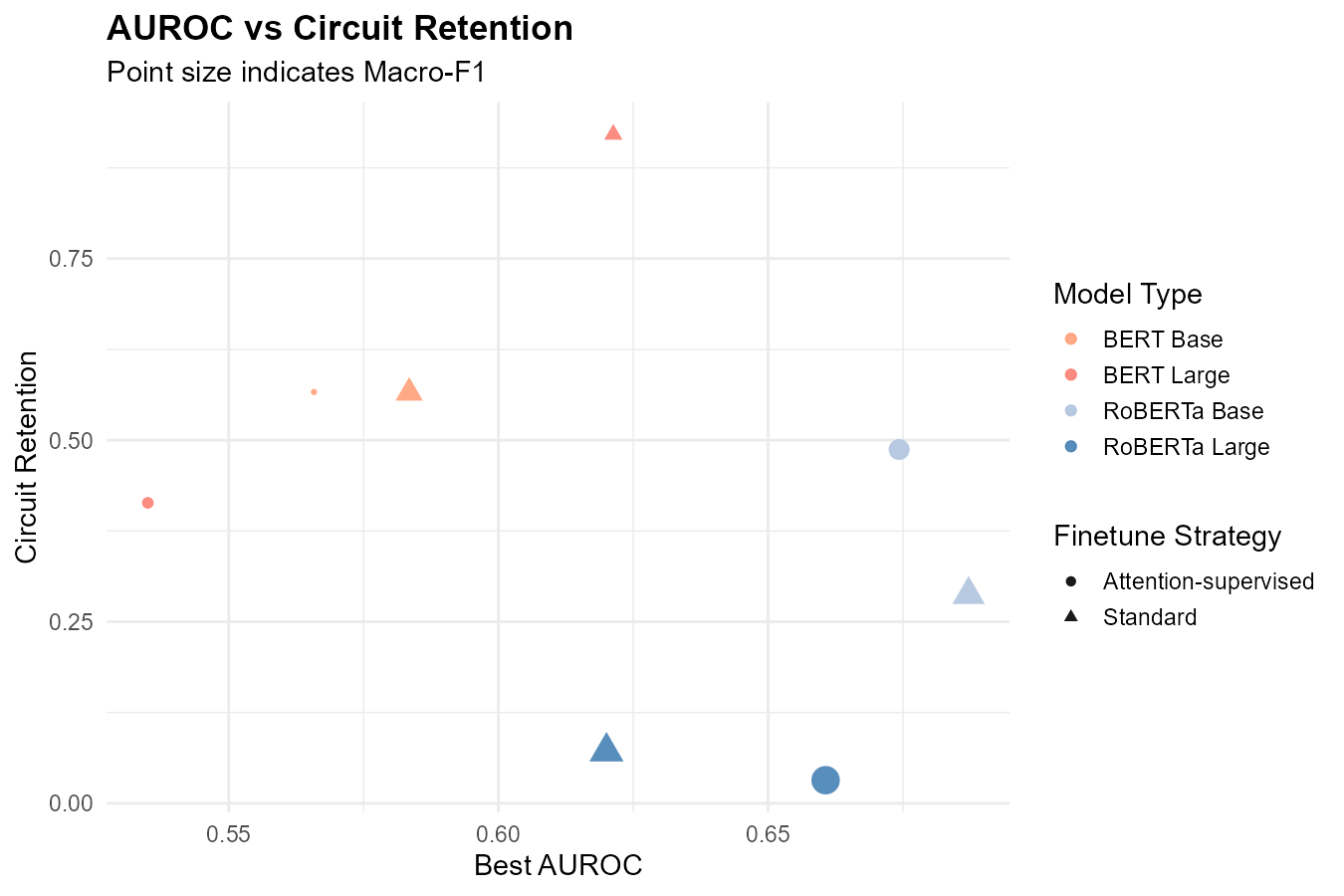}
    \caption{
    Relationship between attribution plausibility (best AUROC across attribution methods)
    and circuit retention ($k=30$) across all model configurations.
    Point size corresponds to Macro F1.
    }
    \label{fig:pareto}
\end{figure}

\paragraph{Relationships among explanation dimensions.}

Figure~\ref{fig:pareto} compares plausibility of attribution and circuit retention in all model configurations. We do not observe a consistent monotonic relationship between the two quantities. Models with strong rationale alignment do not necessarily exhibit high circuit retention, while some configurations with only moderate plausibility scores achieve relatively recoverable circuits. In particular, the RoBERTa models achieve some of the strongest plausibility scores in our experiments but often display comparatively weak circuit retention, whereas several BERT configurations exhibit the opposite pattern.

These observations suggest that the plausibility of the attribution and the mechanistic faithfulness capture complementary but distinct aspects of the model behavior. Models with strong rationale alignment do not necessarily exhibit highly recoverable circuits, indicating that improvements in one explanation metric do not automatically transfer to another.
\section{Conclusion}
\label{sec:conclusion}

We evaluated BERT and RoBERTa classifiers for media bias detection along three axes: predictive performance, plausibility of explanations, and mechanistic faithfulness. Attention-supervised finetuning serves as an intervention for studying explanation behavior, while attribution analyzes reveal only moderate agreement with expert rationales and substantial variation across architectures and explanation methods. Circuit discovery through activation patching further reveals substantial variation in mechanistic recoverability across models. Rescue analysis in particular suggests that bias prediction emerges from coordination among multiple attention heads rather than a single sparse mechanism. Overall, our findings indicate that plausibility and faithfulness capture complementary but distinct aspects of model behavior and thus should be assessed as non-interchangeable notions of explanation quality. Because explanations in media bias detection may inform fairness judgments, journalistic analysis, and model auditing, we argue that trustworthy systems should be evaluated along both token-level plausibility and component-level faithfulness dimensions, each on its own terms.

\section{Limitations}

\paragraph{Dataset scale and diversity.}
All experiments use the BABE dataset, which is relatively small and topically constrained compared to real-world media environments. The observed attention patterns and circuit structures may therefore reflect dataset-specific artifacts rather than general properties of bias detection. Evaluation on larger and more diverse news corpora is needed.

\paragraph{Statistical power.} The number of model configurations evaluated is relatively small, limiting the strength of conclusions regarding relationships among predictive performance, plausibility, and mechanistic faithfulness. Future work should examine a broader range of architectures, datasets, and training interventions.

\paragraph{Approximate circuit discovery.}
Our mechanistic analysis relies on activation patching with top-$k$ component selection, which provides an approximate view of the model's internal computation. The resulting circuits should not be interpreted as the unique or complete mechanism underlying bias detection. Encoder-based transformers distribute information bidirectionally across layers, making compact causal decomposition harder than in autoregressive architectures.

\paragraph{Architectural scope.}
Our experiments are limited to encoder-only models (BERT and RoBERTa). Circuit discovery may behave differently in decoder-only or instruction-tuned language models, where autoregressive token flow introduces different computational structures.

\section{Ethical Considerations}

\paragraph{Dual-use risk.}
Automated bias detection could be misused to craft less detectable biased content rather than to identify and mitigate it. We emphasize that our methods are intended to support transparency in media analysis, not to enable adversarial manipulation of news framing.

\paragraph{Subjectivity of bias labels.}
The BABE dataset relies on expert annotations that reflect the annotators' cultural and political perspectives. Bias is context-dependent and contested; no single annotation scheme captures all forms of framing. Model outputs should be treated as one signal among many rather than as definitive judgments.

\paragraph{Risk of misapplication.}
Deploying bias classifiers without interpretable explanations risks unjustified censorship or editorial suppression. Our work on explainability is motivated in part by this concern: bias predictions should be accompanied by transparent reasoning to support human decision-making rather than automated content moderation.

\bibliography{items}

@inproceedings{arad-etal-2025-findings,
    title = "Findings of the {B}lackbox{NLP} 2025 Shared Task: Localizing Circuits and Causal Variables in Language Models",
    author = "Arad, Dana  and
      Belinkov, Yonatan  and
      Chen, Hanjie  and
      Kim, Najoung  and
      Mohebbi, Hosein  and
      Mueller, Aaron  and
      Sarti, Gabriele  and
      Tutek, Martin",
    editor = "Belinkov, Yonatan  and
      Mueller, Aaron  and
      Kim, Najoung  and
      Mohebbi, Hosein  and
      Chen, Hanjie  and
      Arad, Dana  and
      Sarti, Gabriele",
    booktitle = "Proceedings of the 8th BlackboxNLP Workshop: Analyzing and Interpreting Neural Networks for NLP",
    month = nov,
    year = "2025",
    address = "Suzhou, China",
    publisher = "Association for Computational Linguistics",
    url = "https://aclanthology.org/2025.blackboxnlp-1.32/",
    doi = "10.18653/v1/2025.blackboxnlp-1.32",
    pages = "543--552",
    ISBN = "979-8-89176-346-3"
}

@inproceedings{syed-etal-2024-attribution,
    title = "Attribution Patching Outperforms Automated Circuit Discovery",
    author = "Syed, Aaquib  and
      Rager, Can  and
      Conmy, Arthur",
    editor = "Belinkov, Yonatan  and
      Kim, Najoung  and
      Jumelet, Jaap  and
      Mohebbi, Hosein  and
      Mueller, Aaron  and
      Chen, Hanjie",
    booktitle = "Proceedings of the 7th BlackboxNLP Workshop: Analyzing and Interpreting Neural Networks for NLP",
    month = nov,
    year = "2024",
    address = "Miami, Florida, US",
    publisher = "Association for Computational Linguistics",
    url = "https://aclanthology.org/2024.blackboxnlp-1.25/",
    doi = "10.18653/v1/2024.blackboxnlp-1.25",
    pages = "407--416"
}

@misc{huang2026biggerisntbettercomprehensive,
      title={When Bigger Isn't Better: A Comprehensive Fairness Evaluation of Political Bias in Multi-News Summarisation}, 
      author={Nannan Huang and Iffat Maab and Junichi Yamagishi},
      year={2026},
      eprint={2604.21309},
      archivePrefix={arXiv},
      primaryClass={cs.CL},
      url={https://arxiv.org/abs/2604.21309}, 
}

@misc{hernandes2024llmsleftrightcenter,
      title={LLMs left, right, and center: Assessing GPT's capabilities to label political bias from web domains}, 
      author={Raphael Hernandes and Giulio Corsi},
      year={2024},
      eprint={2407.14344},
      archivePrefix={arXiv},
      primaryClass={cs.CL},
      url={https://arxiv.org/abs/2407.14344}, 
}

@article{SPINDE2021102505,
title = {Automated identification of bias inducing words in news articles using linguistic and context-oriented features},
journal = {Information Processing and Management},
volume = {58},
number = {3},
pages = {102505},
year = {2021},
issn = {0306-4573},
doi = {https://doi.org/10.1016/j.ipm.2021.102505},
url = {https://www.sciencedirect.com/science/article/pii/S0306457321000157},
author = {Timo Spinde and Lada Rudnitckaia and Jelena Mitrović and Felix Hamborg and Michael Granitzer and Bela Gipp and Karsten Donnay}
}

@article{MANDREKAR20101315,
title = {Receiver Operating Characteristic Curve in Diagnostic Test Assessment},
journal = {Journal of Thoracic Oncology},
volume = {5},
number = {9},
pages = {1315-1316},
year = {2010},
issn = {1556-0864},
doi = {https://doi.org/10.1097/JTO.0b013e3181ec173d},
url = {https://www.sciencedirect.com/science/article/pii/S1556086415306043},
author = {Jayawant N. Mandrekar}
}

@misc{olsson2022incontextlearninginductionheads,
      title={In-context Learning and Induction Heads}, 
      author={Catherine Olsson and Nelson Elhage and Neel Nanda and Nicholas Joseph and Nova DasSarma and Tom Henighan and Ben Mann and Amanda Askell and Yuntao Bai and Anna Chen and Tom Conerly and Dawn Drain and Deep Ganguli and Zac Hatfield-Dodds and Danny Hernandez and Scott Johnston and Andy Jones and Jackson Kernion and Liane Lovitt and Kamal Ndousse and Dario Amodei and Tom Brown and Jack Clark and Jared Kaplan and Sam McCandlish and Chris Olah},
      year={2022},
      eprint={2209.11895},
      archivePrefix={arXiv},
      primaryClass={cs.LG},
      url={https://arxiv.org/abs/2209.11895}, 
}

@article{elhage2021mathematical,
   title={A Mathematical Framework for Transformer Circuits},
   author={Elhage, Nelson and Nanda, Neel and Olsson, Catherine and Henighan, Tom and Joseph, Nicholas and Mann, Ben and Askell, Amanda and Bai, Yuntao and Chen, Anna and Conerly, Tom and DasSarma, Nova and Drain, Dawn and Ganguli, Deep and Hatfield-Dodds, Zac and Hernandez, Danny and Jones, Andy and Kernion, Jackson and Lovitt, Liane and Ndousse, Kamal and Amodei, Dario and Brown, Tom and Clark, Jack and Kaplan, Jared and McCandlish, Sam and Olah, Chris},
   year={2021},
   journal={Transformer Circuits Thread},
   url={https://transformer-circuits.pub/2021/framework/index.html}
}

@misc{conmy2023automatedcircuitdiscoverymechanistic,
      title={Towards Automated Circuit Discovery for Mechanistic Interpretability}, 
      author={Arthur Conmy and Augustine N. Mavor-Parker and Aengus Lynch and Stefan Heimersheim and Adrià Garriga-Alonso},
      year={2023},
      eprint={2304.14997},
      archivePrefix={arXiv},
      primaryClass={cs.LG},
      url={https://arxiv.org/abs/2304.14997}, 
}

@inproceedings{cho-etal-2025-mechanistic,
    title = "Mechanistic Fine-tuning for In-context Learning",
    author = "Cho, Hakaze  and
      Luo, Peng  and
      Kato, Mariko  and
      Kaenbyou, Rin  and
      Inoue, Naoya",
    editor = "Belinkov, Yonatan  and
      Mueller, Aaron  and
      Kim, Najoung  and
      Mohebbi, Hosein  and
      Chen, Hanjie  and
      Arad, Dana  and
      Sarti, Gabriele",
    booktitle = "Proceedings of the 8th BlackboxNLP Workshop: Analyzing and Interpreting Neural Networks for NLP",
    month = nov,
    year = "2025",
    address = "Suzhou, China",
    publisher = "Association for Computational Linguistics",
    url = "https://aclanthology.org/2025.blackboxnlp-1.21/",
    doi = "10.18653/v1/2025.blackboxnlp-1.21",
    pages = "330--357",
    ISBN = "979-8-89176-346-3"
}

@misc{clark2019doesbertlookat,
      title={What Does BERT Look At? An Analysis of BERT's Attention}, 
      author={Kevin Clark and Urvashi Khandelwal and Omer Levy and Christopher D. Manning},
      year={2019},
      eprint={1906.04341},
      archivePrefix={arXiv},
      primaryClass={cs.CL},
      url={https://arxiv.org/abs/1906.04341}, 
}

@article{lyu_towards_2024,
	title = {Towards Faithful Model Explanation in {NLP}: A Survey},
	volume = {50},
	url = {https://aclanthology.org/2024.cl-2.6/},
	doi = {10.1162/coli_a_00511},
	shorttitle = {Towards Faithful Model Explanation in {NLP}},
	pages = {657--723},
	number = {2},
	journal = {Computational Linguistics},
	author = {Lyu, Qing and Apidianaki, Marianna and Callison-Burch, Chris},
	urldate = {2025-07-28},
	year = {2024},
	date = {2024-06},
	note = {Place: Cambridge, {MA}
Publisher: {MIT} Press},
}

@inproceedings{mylonas_improving_2022,
	title = {Improving Attention-Based Interpretability of Text Classification Transformers},
	url = {https://www.researchgate.net/publication/363765786},
	doi = {10.1007/978-3-031-17849-8_15},
	booktitle = {Artificial Neural Networks and Machine Learning -- {ICANN} 2022},
	author = {Mylonas, Nikolaos and Mollas, Ioannis and Tsoumakas, Grigorios},
	year = {2022},
}

@incollection{naim_explaining_2024,
	title = {On Explaining with Attention Matrices},
	url = {http://arxiv.org/abs/2410.18541},
	author = {Naim, Omar and Asher, Nicholas},
	urldate = {2025-07-31},
	year = {2024},
	date = {2024-10-16},
	doi = {10.3233/FAIA240594},
	eprinttype = {arxiv},
	eprint = {2410.18541 [cs]},
}

@misc{hao_self-attention_2021,
	title = {Self-Attention Attribution: Interpreting Information Interactions Inside Transformer},
	url = {http://arxiv.org/abs/2004.11207},
	doi = {10.48550/arXiv.2004.11207},
	shorttitle = {Self-Attention Attribution},
	number = {{arXiv}:2004.11207},
	publisher = {{arXiv}},
	author = {Hao, Yaru and Dong, Li and Wei, Furu and Xu, Ke},
	urldate = {2025-07-31},
	year = {2021},
	date = {2021-02-25},
	eprinttype = {arxiv},
	eprint = {2004.11207 [cs]},
}

@misc{wei_chain--thought_2023,
	title = {Chain-of-Thought Prompting Elicits Reasoning in Large Language Models},
	url = {http://arxiv.org/abs/2201.11903},
	doi = {10.48550/arXiv.2201.11903},
	number = {{arXiv}:2201.11903},
	publisher = {{arXiv}},
	author = {Wei, Jason and Wang, Xuezhi and Schuurmans, Dale and Bosma, Maarten and Ichter, Brian and Xia, Fei and Chi, Ed and Le, Quoc and Zhou, Denny},
	urldate = {2025-07-31},
	year = {2023},
	date = {2023-01-10},
	eprinttype = {arxiv},
	eprint = {2201.11903 [cs]},
}

@inproceedings{yeo_how_2024,
	title = {How Interpretable are Reasoning Explanations from Prompting Large Language Models?},
	url = {http://arxiv.org/abs/2402.11863},
	doi = {10.18653/v1/2024.findings-naacl.138},
	pages = {2148--2164},
	booktitle = {Findings of the Association for Computational Linguistics: {NAACL} 2024},
	author = {Yeo, Wei Jie and Satapathy, Ranjan and Goh, Rick Siow Mong and Cambria, Erik},
	urldate = {2025-07-31},
	year = {2024},
	date = {2024},
	eprinttype = {arxiv},
	eprint = {2402.11863 [cs]},
}

@misc{spinde_media_2024,
	title = {The Media Bias Taxonomy: A Systematic Literature Review on the Forms and Automated Detection of Media Bias},
	url = {http://arxiv.org/abs/2312.16148},
	doi = {10.48550/arXiv.2312.16148},
	shorttitle = {The Media Bias Taxonomy},
	number = {{arXiv}:2312.16148},
	publisher = {{arXiv}},
	author = {Spinde, Timo and Hinterreiter, Smi and Haak, Fabian and Ruas, Terry and Giese, Helge and Meuschke, Norman and Gipp, Bela},
	urldate = {2025-08-15},
	year = {2024},
	date = {2024-01-10},
	eprinttype = {arxiv},
	eprint = {2312.16148 [cs]},
}

@article{zheng_attention_2025,
	title = {Attention heads of large language models},
	volume = {6},
	issn = {2666-3899},
	doi = {10.1016/j.patter.2025.101176},
	pages = {101176},
	number = {2},
	journal = {Patterns},
	shortjournal = {Patterns (N Y)},
	author = {Zheng, Zifan and Wang, Yezhaohui and Huang, Yuxin and Song, Shichao and Yang, Mingchuan and Tang, Bo and Xiong, Feiyu and Li, Zhiyu},
	year = {2025},
	date = {2025-02-14},
	pmid = {40041856},
	pmcid = {PMC11873009},
}

@misc{wang_adaptable_2024,
	title = {Adaptable and Reliable Text Classification using Large Language Models},
	url = {http://arxiv.org/abs/2405.10523},
	doi = {10.48550/arXiv.2405.10523},
	number = {{arXiv}:2405.10523},
	publisher = {{arXiv}},
	author = {Wang, Zhiqiang and Pang, Yiran and Lin, Yanbin and Zhu, Xingquan},
	urldate = {2025-08-15},
	year = {2024},
	date = {2024-12-07},
	eprinttype = {arxiv},
	eprint = {2405.10523 [cs]},
	note = {version: 3},
}

@inproceedings{spinde_neural_2021,
	title = {Neural Media Bias Detection Using Distant Supervision With {BABE} -- Bias Annotations By Experts},
	url = {http://arxiv.org/abs/2209.14557},
	doi = {10.18653/v1/2021.findings-emnlp.101},
	pages = {1166--1177},
	booktitle = {Findings of the Association for Computational Linguistics: {EMNLP} 2021},
	author = {Spinde, Timo and Plank, Manuel and Krieger, Jan-David and Ruas, Terry and Gipp, Bela and Aizawa, Akiko},
	urldate = {2025-12-05},
	year = {2021},
	date = {2021},
	eprinttype = {arxiv},
	eprint = {2209.14557 [cs]},
}

@misc{deyoung_eraser_2020,
	title = {{ERASER}: A Benchmark to Evaluate Rationalized {NLP} Models},
	url = {http://arxiv.org/abs/1911.03429},
	doi = {10.48550/arXiv.1911.03429},
	shorttitle = {{ERASER}},
	number = {{arXiv}:1911.03429},
	publisher = {{arXiv}},
	author = {{DeYoung}, Jay and Jain, Sarthak and Rajani, Nazneen Fatema and Lehman, Eric and Xiong, Caiming and Socher, Richard and Wallace, Byron C.},
	urldate = {2025-12-05},
	year = {2020},
	date = {2020-04-24},
	eprinttype = {arxiv},
	eprint = {1911.03429 [cs]},
}

@misc{just_data-centric_2025,
	title = {Data-Centric Human Preference with Rationales for Direct Preference Alignment},
	url = {http://arxiv.org/abs/2407.14477},
	doi = {10.48550/arXiv.2407.14477},
	number = {{arXiv}:2407.14477},
	publisher = {{arXiv}},
	author = {Just, Hoang Anh and Jin, Ming and Sahu, Anit and Phan, Huy and Jia, Ruoxi},
	urldate = {2025-12-05},
	year = {2025},
	date = {2025-07-13},
	eprinttype = {arxiv},
	eprint = {2407.14477 [cs]},
}

@article{tursunalieva_making_2024,
	title = {Making Sense of Machine Learning: A Review of Interpretation Techniques and Their Applications},
	volume = {14},
	issn = {2076-3417},
	url = {https://www.mdpi.com/2076-3417/14/2/496},
	doi = {10.3390/app14020496},
	shorttitle = {Making Sense of Machine Learning},
	pages = {496},
	number = {2},
	journal = {Applied Sciences},
	shortjournal = {Applied Sciences},
	author = {Tursunalieva, Ainura and Alexander, David L. J. and Dunne, Rob and Li, Jiaming and Riera, Luis and Zhao, Yanchang},
	urldate = {2025-12-05},
	year = {2024},
	date = {2024-01-05},
	langid = {english},
}

@misc{luo_understanding_2024,
	title = {From Understanding to Utilization: A Survey on Explainability for Large Language Models},
	url = {http://arxiv.org/abs/2401.12874},
	doi = {10.48550/arXiv.2401.12874},
	shorttitle = {From Understanding to Utilization},
	number = {{arXiv}:2401.12874},
	publisher = {{arXiv}},
	author = {Luo, Haoyan and Specia, Lucia},
	urldate = {2025-12-05},
	year = {2024},
	date = {2024-02-22},
	eprinttype = {arxiv},
	eprint = {2401.12874 [cs]},
}

@misc{wiegreffe_attention_2019,
	title = {Attention is not not Explanation},
	url = {http://arxiv.org/abs/1908.04626},
	doi = {10.48550/arXiv.1908.04626},
	number = {{arXiv}:1908.04626},
	publisher = {{arXiv}},
	author = {Wiegreffe, Sarah and Pinter, Yuval},
	urldate = {2025-12-05},
	year = {2019},
	date = {2019-09-05},
	eprinttype = {arxiv},
	eprint = {1908.04626 [cs]},
}

@article{hamborg_automated_2019,
	title = {Automated identification of media bias in news articles: an interdisciplinary literature review},
	volume = {20},
	issn = {1432-5012, 1432-1300},
	url = {http://link.springer.com/10.1007/s00799-018-0261-y},
	doi = {10.1007/s00799-018-0261-y},
	shorttitle = {Automated identification of media bias in news articles},
	pages = {391--415},
	number = {4},
	journal = {International Journal on Digital Libraries},
	shortjournal = {Int J Digit Libr},
	author = {Hamborg, Felix and Donnay, Karsten and Gipp, Bela},
	urldate = {2025-12-12},
	year = {2019},
	date = {2019-12},
	langid = {english},
}

@misc{devlin2019bertpretrainingdeepbidirectional,
      title={BERT: Pre-training of Deep Bidirectional Transformers for Language Understanding}, 
      author={Jacob Devlin and Ming-Wei Chang and Kenton Lee and Kristina Toutanova},
      year={2019},
      eprint={1810.04805},
      archivePrefix={arXiv},
      primaryClass={cs.CL},
      url={https://arxiv.org/abs/1810.04805}, 
}

@misc{liu2019robertarobustlyoptimizedbert,
      title={RoBERTa: A Robustly Optimized BERT Pretraining Approach}, 
      author={Yinhan Liu and Myle Ott and Naman Goyal and Jingfei Du and Mandar Joshi and Danqi Chen and Omer Levy and Mike Lewis and Luke Zettlemoyer and Veselin Stoyanov},
      year={2019},
      eprint={1907.11692},
      archivePrefix={arXiv},
      primaryClass={cs.CL},
      url={https://arxiv.org/abs/1907.11692}, 
}

@misc{abnar2020quantifyingattentionflowtransformers,
      title={Quantifying Attention Flow in Transformers}, 
      author={Samira Abnar and Willem Zuidema},
      year={2020},
      eprint={2005.00928},
      archivePrefix={arXiv},
      primaryClass={cs.LG},
      url={https://arxiv.org/abs/2005.00928}, 
}

@misc{jain2019attentionexplanation,
      title={Attention is not Explanation}, 
      author={Sarthak Jain and Byron C. Wallace},
      year={2019},
      eprint={1902.10186},
      archivePrefix={arXiv},
      primaryClass={cs.CL},
      url={https://arxiv.org/abs/1902.10186}, 
}

\appendix
\raggedbottom
\renewcommand{\thetable}{\Alph{section}\arabic{table}}
\renewcommand{\thefigure}{\Alph{section}\arabic{figure}}

\section{Licenses and Artifact Usage}
Our experiments use the BABE dataset \citep{spinde_neural_2021}, which is distributed under the CC BY-NC-SA 4.0 license for non-commercial research use. We additionally use pretrained BERT models released under the Apache License 2.0 and RoBERTa models released under the MIT License through the HuggingFace Transformers library. All experiments comply with the respective licenses and intended research usage terms of these artifacts.

\section{Dataset Statistics}
\setcounter{table}{0}
\setcounter{figure}{0}
\label{app:dataset}

Table~\ref{tab:dataset_stats} summarizes the BABE dataset used in our experiments after preprocessing and class balancing. Notably, expert-annotated rationales cover only 6.18\% of tokens on average, with a mean of 1.72 rationale tokens per biased sentence. This extreme sparsity makes plausibility evaluation particularly challenging, as attribution methods must identify a small number of signal-bearing tokens against a large background of uninformative context.

\begin{table}[h]
\centering
\small
\resizebox{\columnwidth}{!}{
\begin{tabular}{lr}
\toprule
\textbf{Statistic} & \textbf{Value} \\
\midrule
Total sentences & 2762 \\
Biased sentences & 1381 \\
Unbiased sentences & 1381 \\
Avg.\ sentence length (tokens) & 31.38 \\
Avg.\ rationale tokens per biased sentence & 1.72 \\
Avg.\ rationale coverage (\%) & 6.18 \\
\midrule
Train / Val / Test split & 70 / 15 / 15 \\
Class balancing & Yes (downsampled) \\
\bottomrule
\end{tabular}
}
\caption{BABE dataset statistics after preprocessing and class balancing.}
\label{tab:dataset_stats}
\end{table}

\section{Hyperparameter Configuration}
\setcounter{table}{0}
\setcounter{figure}{0}
\label{app:hyperparams}

Table~\ref{tab:hyperparams} lists the hyperparameters used across all experiments. All models were trained using the same configuration unless otherwise noted.

\begin{table}[h]
\centering
\small
\begin{tabular}{lr}
\toprule
\textbf{Hyperparameter} & \textbf{Value} \\
\midrule
Optimizer & AdamW \\
Learning rate & $2\times10^{-5}$ \\
Learning rate schedule & Linear decay\\
Batch size & 16 \\
Max sequence length & 128 \\
Number of epochs & 4 \\
Early stopping patience & N/A \\
Attention loss weight $\lambda$ & 0.5 (selected from sweep) \\
Random seeds & 5 \\
\bottomrule
\end{tabular}
\caption{Hyperparameter configuration for all fine-tuning experiments.}
\label{tab:hyperparams}
\end{table}

Table~\ref{tab:lambda_sweep} reports validation-set performance across the full
$\lambda$ sweep used for hyperparameter selection on BERT-base. The results reveal a non-monotonic relationship: moderate values ($\lambda \in [0.2, 0.7]$) improve both Macro F1 and rationale AUROC over the unsupervised baseline ($\lambda = 0$), while stronger supervision ($\lambda \geq 1.5$) degrades both metrics. At $\lambda = 5.0$, Macro F1 drops by over five points, indicating that excessive attention pressure distorts the learned representations and harms the primary classification objective. The selected value of $\lambda = 0.5$ achieves the best joint performance.

\begin{table}[h]
\centering
\small
\begin{tabular}{lcc}
\toprule
\textbf{$\lambda$} & \textbf{Macro F1} & \textbf{AUROC (Grad$\times$Attn)} \\
\midrule
0.0 & 0.818 & 0.623 \\
0.2 & 0.822 & 0.641 \\
0.5 & \textbf{0.824} & \textbf{0.658} \\
0.7 & 0.822 & 0.651 \\
1.0 & 0.817 & 0.644 \\
1.5 & 0.819 & 0.628 \\
2.0 & 0.801 & 0.611 \\
5.0 & 0.764 & 0.573 \\
\bottomrule
\end{tabular}
\caption{
Attention supervision weight ($\lambda$) sweep on the validation set used for
hyperparameter selection.
Moderate supervision strengths improve both classification performance and rationale plausibility, while overly strong supervision degrades task performance and explanation quality. 
$\lambda = 0.5$ provides the best tradeoff between predictive accuracy and rationale alignment.
}
\label{tab:lambda_sweep}
\end{table}

All experiments were executed using a SLURM-managed computing cluster. 
Each run requested a GPU with 32GB of memory, while the specific GPU model varied depending on resource availability across compute nodes. Table~\ref{tab:compute} reports the approximate training time and total compute for each model configuration.

\begin{table}[h]
\centering
\small
\begin{tabular}{lcc}
\toprule
\textbf{Model} & \textbf{Parameters} & \textbf{Train Time} \\
\midrule
BERT-base    & 110M & $\sim$3 min \\
BERT-large   & 340M & $\sim$9 min \\
RoBERTa-base & 125M & $\sim$4 min \\
RoBERTa-large & 355M & $\sim$10 min \\
\midrule
\multicolumn{2}{l}{$\lambda$ sweep (BERT-base only)} & 8 values $\times$ 5 seeds \\
\multicolumn{2}{l}{Total GPU hours (all experiments)} & $\sim$3.5 GPU-hours \\
\bottomrule
\end{tabular}
\caption{Computational budget. Training times are per single run (one seed, one $\lambda$ setting)
on a single GPU. Times are estimated based on dataset size (1{,}933 training
samples), batch size 16, sequence length 128, and 4 epochs.}
\label{tab:compute}
\end{table}
\FloatBarrier

\section{Per-Class and Per-Seed Results}
\setcounter{table}{0}
\setcounter{figure}{0}
\label{app:perclass}

Table~\ref{tab:per_class} disaggregates classification performance by class and finetuning condition, averaged across 5 sampling seeds. Across most architectures, attention-supervised models tend to increase precision on the biased class while reducing recall, suggesting that rationale supervision encourages more conservative bias predictions. For example, BERT-base attention-supervised achieves 0.907 biased precision (vs.\ 0.880 standard) but only 0.732 biased recall (vs.\ 0.789). RoBERTa-base exhibits the opposite behavior, trading a small reduction in biased precision (0.901 to 0.889) for a substantial increase in biased recall (0.803 to 0.841), resulting in improved biased-class F1. Standard deviations remain small across seeds ($\leq 0.038$), indicating stable training dynamics.

\begin{table}[H]
\centering
\small
\setlength{\tabcolsep}{3pt}
\resizebox{\columnwidth}{!}{
\begin{tabular}{ll ccc ccc}
\toprule
\multirow{2}{*}{\textbf{Model}} &
\multirow{2}{*}{\textbf{FT}} &
\multicolumn{3}{c}{\textbf{Biased}} &
\multicolumn{3}{c}{\textbf{Unbiased}} \\
\cmidrule(lr){3-5} \cmidrule(lr){6-8}
& & \textbf{P} & \textbf{R} & \textbf{F1} & \textbf{P} & \textbf{R} & \textbf{F1} \\
\midrule
\multirow{2}{*}{BERT-b}
 & Std  & .880\tss{15} & .789\tss{22} & .832\tss{15} & .766\tss{19} & .865\tss{19} & .813\tss{14} \\
 & Attn & .907\tss{11} & .732\tss{23} & .810\tss{14} & .729\tss{16} & .906\tss{13} & .808\tss{10} \\
\midrule
\multirow{2}{*}{BERT-l}
 & Std  & .868\tss{13} & .801\tss{38} & .832\tss{20} & .773\tss{32} & .846\tss{21} & .808\tss{16} \\
 & Attn & .877\tss{18} & .800\tss{31} & .837\tss{24} & .775\tss{30} & .860\tss{20} & .815\tss{24} \\
\midrule
\multirow{2}{*}{RoB-b}
 & Std  & .901\tss{20} & .803\tss{25} & .849\tss{12} & .783\tss{18} & .888\tss{26} & .832\tss{11} \\
 & Attn & .889\tss{10} & .841\tss{22} & .864\tss{10} & .813\tss{19} & .868\tss{16} & .840\tss{08} \\
\midrule
\multirow{2}{*}{RoB-l}
 & Std  & .908\tss{14} & .831\tss{34} & .868\tss{20} & .810\tss{29} & .894\tss{19} & .849\tss{18} \\
 & Attn & .918\tss{18} & .798\tss{20} & .854\tss{18} & .783\tss{20} & .911\tss{20} & .842\tss{19} \\
\bottomrule
\end{tabular}
}
\caption{Per-class precision (P), recall (R), and F1. Subscripts denote std.\ dev.\ ($\times 10^{-3}$) across 5 seeds. BERT-b/l = BERT base/large; RoB-b/l = RoBERTa base/large; Std = standard; Attn = attention-supervised.}
\label{tab:per_class}
\end{table}

\section{Full Circuit Discovery Results}
\setcounter{table}{0}
\setcounter{figure}{0}
\label{app:circuit}
Table~\ref{tab:acdc_full_results} presents the complete activation patching results across all models, finetuning conditions, and circuit sizes ($k=1$ to $30$). Across nearly all configurations, retention increases with circuit size, indicating that predictive information is distributed across multiple attention heads rather than concentrated in a single component. Rescue scores are typically negative for very small circuits and become positive only as additional heads are incorporated, suggesting that meaningful predictive behavior emerges from coordinated computation across multiple components.

Circuit behavior varies substantially across architectures. BERT-base and RoBERTa-base exhibit steadily increasing retention and rescue as circuit size grows, indicating progressively more recoverable internal computation. RoBERTa-large, by contrast, maintains weak retention and rescue scores even for larger circuits, suggesting more distributed predictive representations. Because retention is normalized by the base predictive gap, we additionally report Rescue as a denominator-free measure of recovered predictive signal.

\begin{table}[H]
\centering
\footnotesize
\setlength{\tabcolsep}{2pt}
\renewcommand{\arraystretch}{0.85}
\resizebox{\columnwidth}{!}{
\begin{tabular}{llcccccc}
\toprule
\multirow{2}{*}{\textbf{Model}} &
\multirow{2}{*}{\textbf{FT}} &
\multirow{2}{*}{\textbf{Base}} &
\multirow{2}{*}{\textbf{$k$}} &
\multirow{2}{*}{\textbf{None}} &
\multirow{2}{*}{\textbf{Circ}} &
\multicolumn{2}{c}{\textbf{Faithfulness}} \\
\cmidrule(lr){7-8}
& & & & & & \textbf{Ret.} & \textbf{Resc.} \\
\midrule

\multirow{12}{*}{BERT-b}
& \multirow{6}{*}{Std}
& \multirow{6}{*}{0.3059}
& 1  & 0.3045 & 0.0560 & 0.1830 & $-$0.2485 \\
& & & 3  & 0.2958 & 0.0633 & 0.2071 & $-$0.2325 \\
& & & 5  & 0.2712 & 0.0778 & 0.2542 & $-$0.1935 \\
& & & 10 & 0.2476 & 0.0887 & 0.2900 & $-$0.1589 \\
& & & 20 & 0.1748 & 0.1387 & 0.4533 & $-$0.0362 \\
& & & 30 & 0.0825 & 0.1729 & 0.5653 & 0.0905 \\
\cmidrule(lr){2-8}

& \multirow{6}{*}{Attn}
& \multirow{6}{*}{0.1200}
& 1  & 0.0990 & 0.0009 & 0.0076 & $-$0.0981 \\
& & & 3  & 0.0885 & 0.0122 & 0.1015 & $-$0.0763 \\
& & & 5  & 0.0667 & 0.0149 & 0.1238 & $-$0.0518 \\
& & & 10 & 0.0560 & 0.0285 & 0.2372 & $-$0.0275 \\
& & & 20 & 0.0379 & 0.0445 & 0.3710 & 0.0067 \\
& & & 30 & 0.0300 & 0.0680 & 0.5663 & 0.0380 \\
\midrule

\multirow{12}{*}{BERT-l}
& \multirow{6}{*}{Std}
& \multirow{6}{*}{0.0163}
& 1  & 0.0177 & $-$0.0001 & $-$0.0061 & $-$0.0178 \\
& & & 3  & 0.0165 & 0.0007 & 0.0415 & $-$0.0158 \\
& & & 5  & 0.0154 & 0.0015 & 0.0911 & $-$0.0139 \\
& & & 10 & 0.0234 & 0.0048 & 0.2922 & $-$0.0187 \\
& & & 20 & 0.0199 & 0.0094 & 0.5754 & $-$0.0106 \\
& & & 30 & 0.0159 & 0.0150 & 0.9210 & $-$0.0009 \\
\cmidrule(lr){2-8}

& \multirow{6}{*}{Attn}
& \multirow{6}{*}{0.0627}
& 1  & 0.0564 & $-$0.0023 & $-$0.0371 & $-$0.0588 \\
& & & 3  & 0.0481 & $-$0.0037 & $-$0.0588 & $-$0.0518 \\
& & & 5  & 0.0412 & $-$0.0009 & $-$0.0141 & $-$0.0421 \\
& & & 10 & 0.0272 & 0.0052 & 0.0836 & $-$0.0219 \\
& & & 20 & 0.0234 & 0.0138 & 0.2202 & $-$0.0096 \\
& & & 30 & 0.0201 & 0.0260 & 0.4137 & 0.0058 \\
\midrule

\multirow{12}{*}{RoB-b}
& \multirow{6}{*}{Std}
& \multirow{6}{*}{0.1259}
& 1  & 0.1040 & $-$0.0006 & $-$0.0045 & $-$0.1046 \\
& & & 3  & 0.0876 & $-$0.0011 & $-$0.0086 & $-$0.0886 \\
& & & 5  & 0.0495 & $-$0.0001 & $-$0.0011 & $-$0.0497 \\
& & & 10 & 0.0088 & 0.0004 & 0.0035 & $-$0.0083 \\
& & & 20 & 0.0049 & 0.0116 & 0.0918 & 0.0066 \\
& & & 30 & 0.0014 & 0.0362 & 0.2873 & 0.0347 \\
\cmidrule(lr){2-8}

& \multirow{6}{*}{Attn}
& \multirow{6}{*}{0.0869}
& 1  & 0.0638 & $-$0.0000 & 0.0000 & $-$0.0638 \\
& & & 3  & 0.0418 & 0.0029 & 0.0338 & $-$0.0388 \\
& & & 5  & 0.0230 & 0.0063 & 0.0721 & $-$0.0167 \\
& & & 10 & 0.0064 & 0.0164 & 0.1882 & 0.0100 \\
& & & 20 & $-$0.0028 & 0.0213 & 0.2446 & 0.0240 \\
& & & 30 & $-$0.0037 & 0.0423 & 0.4871 & 0.0460 \\
\midrule

\multirow{12}{*}{RoB-l}
& \multirow{6}{*}{Std}
& \multirow{6}{*}{0.0468}
& 1  & 0.0429 & $-$0.0001 & $-$0.0024 & $-$0.0430 \\
& & & 3  & 0.0095 & 0.0009 & 0.0183 & $-$0.0086 \\
& & & 5  & 0.0110 & 0.0011 & 0.0244 & $-$0.0098 \\
& & & 10 & $-$0.0104 & 0.0009 & 0.0195 & 0.0113 \\
& & & 20 & $-$0.0110 & 0.0021 & 0.0445 & 0.0131 \\
& & & 30 & $-$0.0156 & 0.0033 & 0.0712 & 0.0189 \\
\cmidrule(lr){2-8}

& \multirow{6}{*}{Attn}
& \multirow{6}{*}{0.0707}
& 1  & 0.0698 & $-$0.0002 & $-$0.0030 & $-$0.0700 \\
& & & 3  & 0.0402 & $-$0.0009 & $-$0.0131 & $-$0.0411 \\
& & & 5  & 0.0524 & 0.0000 & 0.0004 & $-$0.0524 \\
& & & 10 & 0.0487 & 0.0004 & 0.0054 & $-$0.0483 \\
& & & 20 & 0.0461 & $-$0.0004 & $-$0.0050 & $-$0.0465 \\
& & & 30 & 0.0374 & 0.0022 & 0.0318 & $-$0.0351 \\
\bottomrule
\end{tabular}
}

\caption{Full circuit discovery results via activation patching. \textbf{Base}: original logit gap; \textbf{None}: gap after ablating the circuit; \textbf{Circ}: gap using only top-$k$ components. \textbf{Ret.}~$= \text{Circ}/|\text{Base}|$ (sufficiency); \textbf{Resc.}~$= \text{Circ} - \text{None}$ (recovered signal). All values rounded to 4 decimal places.}
\label{tab:acdc_full_results}

\end{table}

\end{document}